\documentclass{article}
\PassOptionsToPackage{numbers}{natbib}

\usepackage[main, final]{neurips_2026}

\usepackage[utf8]{inputenc} 
\usepackage[T1]{fontenc}    
\usepackage{hyperref}       
\usepackage{url}            
\usepackage{booktabs}       
\usepackage{amsfonts}       
\usepackage{nicefrac}       
\usepackage{microtype}      
\usepackage{xcolor}         

\usepackage{amsmath}
\usepackage{mathtools}
\usepackage{multirow}
\usepackage{pgfplots}
\usepackage{wrapfig}
\usepackage{subcaption}
\usepackage{enumitem}
\usepackage{amssymb}
\usepackage{amsthm}
\usepackage{graphicx}
\usepackage{algorithm}
\usepackage{algorithmic}

\newcommand{\eps}{\epsilon}

\usepackage[capitalize,noabbrev]{cleveref}

\theoremstyle{plain}

\theoremstyle{definition}

\theoremstyle{remark}

\title{Closing the Train-Test Gap in World Models for Gradient-Based Planning}

\author{%
  Rohun Agrawal\thanks{Equal contribution.} \\
  Columbia University \\
  \And
  Nimit Kalra$^*$ \\
  Columbia University \\
  \And
  Arjun Parthasarathy$^*$ \\
  Columbia University \\
  \And
  Yann LeCun \\
  New York University \\
  \And
  Oumayma Bounou \\
  New York University \\
  \And
  Pavel Izmailov \\
  New York University \\
  \And
  Micah Goldblum \\
  Columbia University \\
}

\begin{document}

\maketitle

\begin{abstract}
  World models paired with model predictive control (MPC) can be trained offline on large-scale datasets of expert trajectories and enable generalization to a wide range of tasks chosen at inference time. Compared to traditional MPC procedures, which rely either on slow search algorithms or on iteratively solving optimization problems exactly, gradient-based planning offers a computationally efficient alternative. However, the performance of gradient-based planning has thus far lagged behind that of other approaches. In this paper, we propose improved methods for training world models that enable efficient gradient-based planning. We begin with the observation that although a world model is trained on a next-state prediction objective, it is used at test-time to instead estimate a sequence of actions. The goal of our work is to close this train-test gap. To that end, we propose train-time data synthesis techniques that enable significantly improved gradient-based planning with existing world models. At test time, our approach outperforms or matches the classical gradient-free cross-entropy method (CEM) across a variety of object manipulation and navigation tasks in 10\% of the time budget.
\end{abstract}

\section{Introduction}

In robotic tasks, anticipating how the actions of an agent affect the state of its environment is fundamental for both prediction~\citep{Finn2016UnsupervisedLFA} and planning \citep{mohanan2018survey, kavraki2002probabilistic}. 
Classical approaches derive models of the environment evolution analytically from first principles, relying on prior knowledge of the environment, the agent, and any uncertainty \citep{goldstein1950classical, siciliano2009robotics,spong2020robot}. In contrast, learning-based methods infer such models directly from data, enabling them to capture complex dynamics and thus improve generalization and robustness to uncertainty \citep{ sutton1998reinforcement, schrittwieser2020mastering, lecun2022path}.

World models \citep{ha2018world}, in particular, have emerged as a powerful paradigm.
Given the current state and an action, the world model predicts the resulting next state.  These models can be learned either from exact state information \citep{sutton1991dyna} or from high-dimensional sensory inputs such as images
\citep{hafner2023mastering}. The latter setup is especially compelling as it enables perception, prediction, and control directly from raw images by leveraging pre-trained visual representations, and removes the need for measuring the precise environment states, which is difficult in practice \citep{assran2023self,Bardes2024RevisitingFPA}. Recently, world models and their predictive capabilities have been leveraged for planning, enabling agents to solve a variety of tasks \citep{hafner2019dream, hafner2019learning, schrittwieser2020mastering, hafner2023mastering,zhou2025dinowmworldmodelspretrained}.
A model of the dynamics is learned offline, while the planning task is defined at inference as a constrained optimization problem: given the current state, find a sequence of actions that results in a state as close as possible to the target state. This inference-time optimization provides an effective alternative to reinforcement learning approaches \citep{sutton1998reinforcement} that often suffer from poor sample efficiency.


\begin{figure*}[!t]
    \centering
    \includegraphics[width=\linewidth]{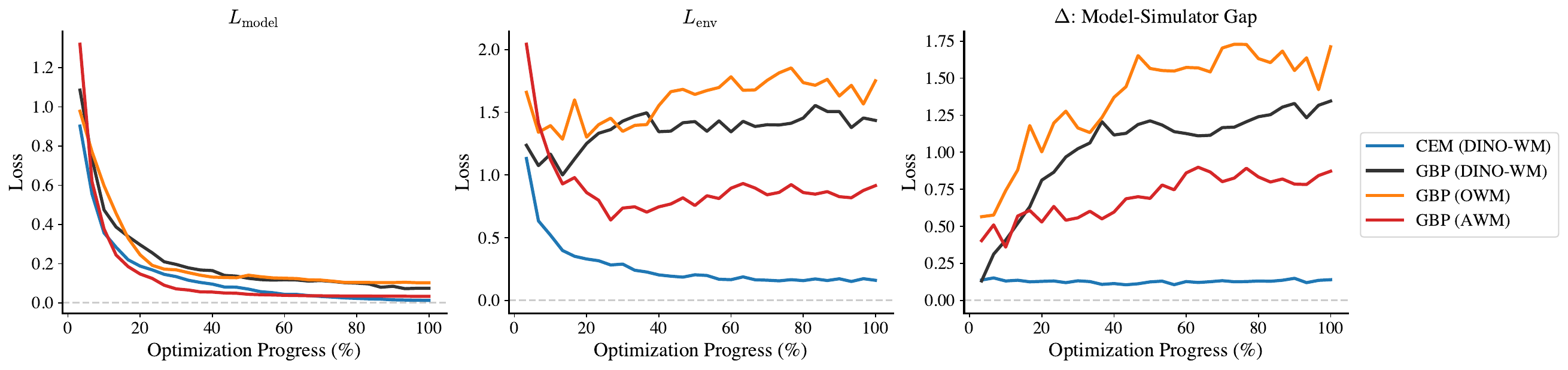}
    \caption{\textbf{Model-Simulator Gap.} We analyze the difference between the planning objective and the true simulator distance as a way of diagnosing the failure modes of GBP. In contrast to CEM that inadvertently optimizes simulator distance along with its planning objective, we see GBP's Model-Simulator Gap grow over the course of optimization. We use 50 unseen trajectories on PointMaze to measure these losses. }
    \label{fig:surrogate_gap}
    \vspace{-15pt}
\end{figure*}

World models are compatible with many model-based planning algorithms. The most commonly used planning methods are search-based, such as the Cross Entropy Method (CEM)~\citep{rubinstein2004cross} and Model Predictive Path Integral control (MPPI)~\citep{williams2017model}, because they have proven effective in practice. However, they are computationally intensive as they require iteratively sampling candidate solutions and performing world model rollouts to evaluate each one, a procedure that scales poorly in high-dimensional spaces.
Gradient-based methods \citep{sv2023gradient}, in contrast, avoid the limitations of sampling by directly exploiting the differentiability of world models to optimize actions end-to-end. These methods eliminate the costly rollouts required by search-based approaches, thus scaling more efficiently in high-dimensional spaces. Despite this promise, gradient-based approaches have thus far seen limited empirical success.

Gradient-based planning (GBP) suffers from a fundamental train-test gap.
World models are typically trained using a next-state prediction objective on datasets of expert trajectories. At test time, however, they are used to optimize a planning objective over sequences of actions. We argue that this mismatch underlies the poor empirical performance of GBP, and we offer two hypotheses to explain why. 1) During planning, the intermediate sequence of actions
explored by gradient descent drives the world model into
states that were not encountered during training. In these
out-of-distribution states, model errors compound, making
the world model unreliable as a surrogate for optimization.
2) The action-level optimization landscape induced by the
world model may be difficult to traverse, containing many
poor local minima or flat regions, which hinders effective
gradient-based optimization.

In this work, we address both of these challenges with two strategies: Online World Modeling and Adversarial World Modeling. Both expand the region of familiar
latent states by continuously adding new trajectories to the
dataset and finetuning the world model on them. To manage
the distribution shift between offline expert trajectories and
predicted trajectories from planning, Online World Modeling uses the environment simulator to correct states along
a trajectory produced by performing GBP. Finetuning on
these corrected trajectories ensures that the world model performs sufficiently well when GBP enters regimes of latent
state space outside of the expert trajectory distribution. To
overcome the difficulties of optimizing over a non-smooth
loss surface during GBP, Adversarial World Modeling perturbs expert trajectories in the direction that maximizes the
world model’s loss. Adversarial finetuning smooths the induced action loss landscape, making it easier to optimize
via gradient-based planning.

Our contributions are summarized as follows:

\begin{enumerate}
    \item We identify the train-test gap in world models used for gradient-based planning, using the model-simulator gap as a diagnostic metric to explain why GBP exploits poor world model predictions. 
    \item  We identify that DAgger-style finetuning and adversarial training are strong candidates to address issues specific to GBP, and adapt them using Online World Modeling (OWM) and Adversarial World Modeling (AWM).

    \item We demonstrate empirically that AWM enables GBP to match or exceed the performance of the gradient-free Cross-Entropy Method across six simulated robotic/navigation tasks and a real-world high-dimensional egocentric pose planning benchmark, while operating in 10× less wall-clock time and establishing a new Pareto frontier for planning efficiency with world models.

    \item We analyze aspects of GBP's failure modes to uncover why OWM is unable to improve GBP performance, revealing that data scale alone is not sufficient. Furthermore, this analysis provides evidence for why AWM significantly improves GBP performance on a variety of domains.
\end{enumerate}

\label{sec:motivation}
\section{Why Gradient-Based Planning Fails}
\subsection{Preliminaries}
\label{sec:pb_formulation}
World models learn environment dynamics by predicting the state resulting from taking an action in the current state. Then, at test time, the learned world model enables planning by simulating future trajectories and guiding action optimization. Formally, a world model approximates the (potentially unknown) dynamics function $h{\,:\,} \mathcal{S} \times \mathcal{A} \to \mathcal{S}$, where $\mathcal{S}$ denotes the state space and $\mathcal{A}$ the action space. The environment evolves according to $s_{t+1} = h(s_t, a_t)$ where $s_t \in \mathcal{S}, a_t \in \mathcal{A}$ denote the state and action at time $t$, respectively.

\paragraph{\textbf{Latent world models.}} In practice, we typically do not have access to the exact state of the environment; instead, we only receive partial observations of it, such as images. In order for a world model to efficiently learn in the high-dimensional observation space $\mathcal{O}$, an embedding function $\Phi_\mu{\,:\, } \mathcal{O} \to \mathcal{Z}$ is employed to map observations to a lower-dimensional latent space $\mathcal{Z}$. Then, given an embedding function $\Phi_\mu$, our goal is to learn a latent world model $f_\theta{\,:\,} \mathcal{Z} \times \mathcal{A} \to \mathcal{Z}$, such that
\begin{equation}
    z_t = \Phi_\mu(o_t), \quad z_{t+1} = f_\theta(z_t, a_t),  \quad \text{for all } t.
\end{equation}
The choice of $\Phi_\mu$ directly affects the expressivity of the latent world model. In this work, we use a fixed encoder pretrained with self-supervised learning that yields rich feature representations out of the box. 

\paragraph{\textbf{Training.}} To train a latent world model, we sample triplets of the form $(o_t, a_t, o_{t + 1})$ from an offline dataset of trajectories $\mathcal{T}$ and minimize the $\ell_2$ distance between the true next latent state $z_{t + 1} = \Phi_\mu(o_{t + 1})$ and the predicted next latent state $\hat{z}_{t + 1}$. This procedure is represented by the following teacher-forcing objective:
\begin{equation}
\min_{\theta}\mathbb{E}_{(o_t, a_t, o_{t + 1}) \sim \mathcal{T}}  \lVert f_\theta(\Phi_{\mu}(o_t), a_t) - \Phi_{\mu}(o_{t + 1})\rVert_2^2.
\end{equation}
Notably, we only minimize this objective with respect to the world model's parameters $\theta$, not those of the potentially large embedding function.

\paragraph{\textbf{Planning.}} At test-time, we use a learned world model to optimize candidate action sequences for reaching a goal state. By recursively applying the world model over an action sequence starting from an initial latent state, we obtain a predicted latent goal state and therefore the distance to the true goal state in latent space. This allows us to find the optimal action sequence
\begin{equation}
\{\hat{a}_t^*\}^H_{t = 1} = \arg \min_{\{\hat{a}_t\}}  \lVert\hat{z}_{H+1} - z_{\text{goal}}\rVert^2_2
\label{eq:plan-objective}
\end{equation}
where $\hat{z}_{H+1}$ is produced by the recursive procedure 
\begin{equation}
    \hat{z}_2 = f_\theta(z_1, \hat{a}_1), \quad 
\hat{z}_{t+1} = f_\theta(\hat{z}_t, \hat{a}_t) \quad \text{for} \quad t>1.
\end{equation}
We use the function $\text{rollout}_f{\,:\,} \mathcal{Z} \times \mathcal{A}^H \to \mathcal{Z}^{H}$ to denote this recursive procedure.

Gradient-based planning (GBP) solves the planning objective (\ref{eq:plan-objective}) via gradient descent. Crucially, since the world model is differentiable, $\nabla_{\{\hat{a}_t\}} \hat{z}_{H + 1} = \nabla_{\{\hat{a}_t\}} \text{rollout}_{f}(z_1, \{\hat{a_t}\})_{H + 1}$ is well-defined. In contrast, the search-based CEM is gradient-free, but requires evaluating substantially more action sequences. We detail GBP in \Cref{algo:gbp} and CEM in \Cref{sec:cem_description}.

 As errors can propagate over long horizons, Model Predictive Control (MPC) is commonly used to repeatedly re-plan by optimizing an $H$-step action sequence but executing only the first $K \le H$ actions before replanning from the updated state.

\begin{algorithm}[tb]
\caption{Gradient-Based Planning via Gradient Descent}
\label{algo:gbp}
\begin{algorithmic}
\STATE {\bfseries Input:} Start state $z_1$, goal state $z_\text{goal}$, world model $f_{\theta}$, horizon $H$, optimization iterations $N$
\STATE Initialize action prediction $\{\hat{a}_t\}_{t=1}^H \sim \mathcal{N}(0, I_H)$
\FOR{$i = 1, \dots, N$}
  \STATE $\hat{z}_{H+1} \leftarrow \text{rollout}_{f}(z_1, \{\hat{a}_t\})_{H + 1}$
  \STATE $\mathcal{L}_{\text{goal}} \leftarrow \lVert\hat{z}_{H+1} - z_{\text{goal}}\rVert^2_2$
  \STATE $\{\hat{a}_t\} \leftarrow \{\hat{a}_t\} - \eta \cdot \nabla_{\{\hat{a}_t\}} \mathcal{L}_{\text{goal}}$
\ENDFOR
\end{algorithmic}
\end{algorithm}

\subsection{Gradient-Based Planning's Failure Modes}
\label{sec:failure_modes}

Gradient-based planning minimizes a planning loss computed through the world model's predictions. Wherever the world model is imperfect, this loss does not reflect what would actually happen in the environment. We find that GBP exploits exactly these inaccuracies, converging to action sequences that achieve low planning loss in the model but fail in the real environment. To measure this, we define the \textit{model-simulator gap} $\Delta = \mathcal{L}_{\text{env}} - \mathcal{L}_{\text{model}}$, where $L_{\text{model}}$ is the planning loss from Algorithm~\ref{algo:gbp} and $\mathcal{L}_{\text{env}} = \|\Phi(s_{\text{final}}) - z_{\text{goal}}\|^2$ is the true distance to the goal obtained by encoding the simulator's final state with the visual encoder $\Phi$. A large positive gap means the world model believes the plan is better than it actually is. In Figure \ref{fig:surrogate_gap}, we see that although GBP appears to optimize $\mathcal{L}_{\text{model}}$ well, it completely fails to optimize $\mathcal{L}_{\text{env}}$, resulting in a large model-simulator gap. While GBP drives the action sequence into regions where the world model diverges from reality, CEM's sampling-based optimization appears more robust to model error, resulting in practically no model-simulator gap.


We hypothesize two reasons why this gap emerges specifically in GBP. First, because GBP is a first-order method, it is more likely to smoothly reach latent states outside the training distribution. In these latent states, the model-simulator gap can be large. Second, the action optimization itself could be unstable, resulting in certain local minima where the model-simulator gap could also be large. This instability likely arises from the fact that the world modeling objective does not encourage smooth input gradients, yet the world model is expected to effectively utilize gradients with respect to its input actions. We now present two finetuning methods designed to improve on these fronts.
\section{Online and Adversarial World Modeling}

\subsection{Online World Modeling}
\label{sec:online_wm}

During gradient-based planning, the action sequences being optimized are not constrained to lie within the distribution of behavior seen during training. World models are typically trained on fixed datasets of expert trajectories, whereas GBP selects actions solely to improve the planning objective, without regard to whether those actions resemble expert behavior. As a result, the optimization process often proposes action sequences that are out of distribution. Optimizing through learned models under such conditions is known to induce adversarial inputs \citep{szegedy2013intriguing, goodfellow2014explaining}. In our setting, these adversarial action sequences drive the world model into regions of the latent state space that were rarely or never observed during training, causing large prediction errors. Even when errors are initially small, they accumulate as the planner rolls the model forward, ultimately degrading long-horizon planning performance.

A natural first approach to closing this gap is to directly expose the world model to the trajectories it encounters during planning. We call this \textbf{Online World Modeling}, which iteratively corrects the trajectories produced by GBP and finetunes the world model on the resulting rollouts. Rather than training solely on expert demonstrations, we repeatedly incorporate trajectories induced by the planner itself, thereby expanding the region of latent states that the world model can reliably predict.

\begin{algorithm}[tb]
\caption{Online World Modeling (OWM)}
\label{algo:online_wm}
\begin{algorithmic}
\STATE {\bfseries Input:} Pretrained world model $f_\theta$, simulator dynamics function $h$, encoder $\Phi_\mu$, dataset of trajectories $\mathcal{T}$, online iterations $N$, horizon $H$, planning optimization iterations $M$
\STATE Initialize new trajectory dataset $\mathcal{T}'$
\FOR{$i = 1, \dots, N$}
    \STATE $\tau_i = (z_1, a_1, z_2, a_2, \dots, a_H, z_{H + 1}) \sim \mathcal{T}$
    \STATE $\{\hat{a}_t \}_{t = 1}^H \gets \text{GBP}(z_{1}, z_{H + 1}, f_\theta, H, M)$
    \STATE $\{s_t'\}_{t = 2}^{H + 1} \gets \text{rollout}_{h}(s_1, \{\hat{a}_t \}) $
    \STATE $\{z_t'\}_{t = 2}^{H + 1}  \gets \{\Phi_\mu(s_t')\}_{t = 2}^{H + 1}$
    \STATE $\tau_i' \gets (z_1, \hat{a}_1, z'_2, \hat{a}_2, \dots, \hat{a}_H, z'_{H+1})$
    \STATE $\mathcal{T}' \gets \mathcal{T}' \cup \tau'_i$
    \STATE Train $f_\theta$ on next-state prediction using $\mathcal{T}'$
\ENDFOR
\end{algorithmic}
\end{algorithm}

First, we conduct GBP using the initial and goal latent states of an expert trajectory $\tau$, yielding a sequence of predicted actions $\{\hat{a}_t\}_{t = 1}^H$.
These actions might send the world model into regions of the latent space that lie outside of the training distribution.
To adjust for this, we obtain a \textit{corrected trajectory}: the actual sequence of states that would result by executing the action sequence $\{\hat{a}_t\}_{t = 1}^H$ in the environment using the true dynamics simulator $h$. We add the corrected trajectory,
\begin{equation}
    \tau'= (z_1, \hat{a}_1, z_2', \hat{a}_2, \dots, z'_{H + 1}),
\end{equation}
to the dataset that the world model trains with every time the dataset is updated.
Re-training on these corrected trajectories expands the training distribution to cover the regions of latent space induced by gradient-based planning, mitigating compounding prediction errors during planning.
We provide more detail in \Cref{algo:online_wm}.

This procedure is a direct adaptation of DAgger (Dataset Aggregation) \citep{ross2011reduction} to the world model setting. Just as DAgger iteratively retrains a policy on its own rollouts corrected by an expert, we retrain the world model on trajectories corrected by the ground-truth simulator. While intuitive, Online World Modeling requires access to the environment simulator during training, which is a major drawback that prevents its use in realistic settings.

\subsection{Adversarial World Modeling}
\label{sec:adversarial_wm}
\begin{figure}[!t]

    \centering
    \includegraphics[width=1\linewidth]{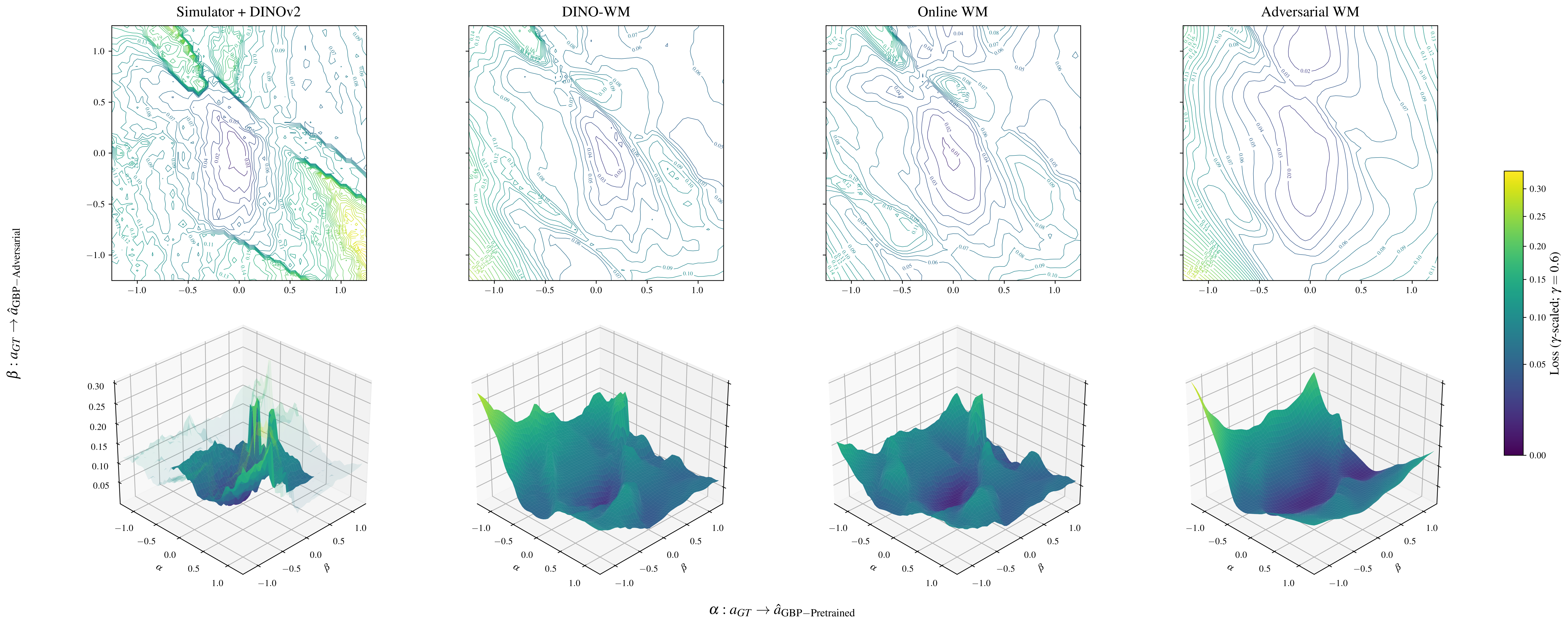}
    \caption{\textbf{Comparison of optimization landscapes.} We compare the actions optimization landscape of DINO-WM \citep{zhou2025dinowmworldmodelspretrained} before and after finetuning with AWM and OWM on the PushT task. AWM yields a smoother landscape with a broader basin around the optimum, while OWM makes a negligible difference in the smoothness of the landscape. Additional details are in \Cref{sec:visualization}.}
    \label{fig:landscape}
\end{figure}

Since world models are only trained on the next-state prediction objective, there is no particular reason for their input gradients to be well-behaved.
Adversarial training has been shown to result in better behaved input gradients \citep{mejia2019robust}, consequently smoothing the input loss surface.
Motivated by this observation, we propose an adversarial training objective that explicitly targets regions of the state-action space where the world model is expected to perform poorly.
These adversarial samples may lie outside the expert trajectory distribution, which can expose the model to precisely the regions that matter for action optimization.
We find that this procedure, which we call \textbf{Adversarial World Modeling}, does in fact smooth the
loss surface of the planning objective (see Figure \ref{fig:landscape}), improving the stability of action-sequence optimization.

Adversarial training improves model robustness by optimizing performance under worst-case perturbations \citep{madry2019deeplearningmodelsresistant}.
An adversarial example is generated by applying a perturbation $\delta$ to an input that maximally increases the model's loss. To train a world model on adversarial examples, we use the objective
\begin{equation}
\label{eq:adversarialobjective}
\begin{split}
\min_\theta \mathbb{E}_{(o_t, a_t, o_{t + 1}) \sim \mathcal{T}} \Bigl[ \max_{\delta_a \in \mathcal{B}_a, \delta_z \in \mathcal{B}_z} & \| f_\theta(\Phi_\mu(o_t) + \delta_z, a_t + \delta_a) - \Phi_\mu(o_{t + 1}) \|_2^2 \Bigr],
\end{split}
\end{equation}
where $\mathcal{B}_a= \{\delta_a: \lVert \delta_a\rVert_{\infty} \leq \epsilon_a \}$ and $\mathcal{B}_z= \{\delta_z: \lVert \delta_z\rVert_{\infty} \leq \epsilon_z \}$ constrain the magnitude of perturbations for given $\epsilon_a, \epsilon_z$. Training on these adversarially perturbed trajectories provides an alternative method to Online World Modeling for surfacing states that may be encountered during planning, without relying on GBP rollouts. This is a significant advantage in settings where simulation is expensive or infeasible. 

We generate adversarial latent states using the Fast Gradient Sign Method (FGSM) \citep{goodfellow2014explaining}, which efficiently approximates the worst-case perturbations that maximize prediction error \citep{fastbetterthanfree}. Although stronger iterative attacks such as Projected Gradient Descent (PGD) can be used, we find that FGSM delivers comparable improvements in GBP performance while being significantly more computationally efficient (see \Cref{sec:fgsm-vs-pgd}). This enables us to generate adversarial samples over entire large-scale offline imitation learning datasets.

For each state-action pair in a given minibatch, we look for small changes to the latent state or action that most increase the world model's prediction error. Let $\epsilon_a, \epsilon_z$ denote the radius of the perturbation to the actions $\{a_t\}$ and latent states $\{z_t\}$ respectively. We compute gradients $\nabla_{\delta_a, \delta_z} \lVert f_\theta(z_t + \delta_z, a_t + \delta_a) - z_{t + 1}\rVert_2^2$ with respect to the perturbations and take a signed gradient ascent step (i.e., in a direction that degrades the prediction) with step sizes $\alpha_a = 1.25 \epsilon_a, \alpha_z = 1.25\epsilon_z$. We clip the result so that each entry of the perturbation stays within the radius. This procedure corresponds to a single step of a PGD-style attack, producing perturbations that lie on the edge of the allowed region where they are maximally challenging for the model. See \Cref{algo:adv_wm} for a detailed treatment.

To initialize the perturbation radii $\epsilon_a, \epsilon_z$, we use scaling factors $\lambda_a, \lambda_z$ and find that Adversarial World Modeling is robust for $0 \le \lambda_a \le 1$ and $0 \le \lambda_z \le 0.5$. Furthermore, we find that fixing $\epsilon_a, \epsilon_z$ to the standard deviation of the initial minibatch is stable across all experiments. Updating this estimate for each batch as in \Cref{algo:adv_wm} yields no consistent improvement in final planning performance. We further analyze design ablations in \Cref{sec:awm-design}.

\section{Experiments}

\subsection{Experimental Setup}
We evaluate on seven tasks spanning a range of planning domains. The six simulated environments include two navigation tasks (PointMaze, Wall), one rigid-body manipulation task (PushT), one continuous control task (Reacher), and two deformable-object manipulation tasks (Rope, Granular). As a real-world evaluation, we additionally evaluate on egocentric pose planning using the Nymeria dataset \citep{nymeria24}, where the agent plans a sequence of joint angle and translation actions to reach a target body pose from egocentric visual observations. Full task descriptions and dataset statistics are provided in Appendix \Cref{sec:task_details}.

We use DINO-WM~\citep{zhou2025dinowmworldmodelspretrained} as our base world model across all tasks so that our experiments emulate planning within the latent space of a large-scale world model. The encoder is a frozen DINOv2~\citep{oquab2024dinov2learningrobustvisual} 
backbone and the transition model $f_\theta$ is implemented as a 
ViT~\citep{dosovitskiy2021imageworth16x16words}.

All planning results are reported under MPC, where an $H$-step action 
sequence is optimized but only the first $K$ actions are kept before replanning 
from the updated state. For GBP, we use Adam~\citep{Kingma2014AdamAM} throughout. To facilitate progress towards the goal during GBP, we set $\mathcal{L}_\text{goal}$ in \Cref{algo:gbp} to an alternate loss function: Weighted Goal Loss (WGL). Instead of the standard goal loss function that only minimizes the $\ell_2$-distance between the final latent state produced by planning actions and the goal latent state, WGL encourages intermediate latent states to also be close to the goal latent state. WGL is detailed in \Cref{sec:wgl}. 

We compare the planning performance of DINO-WM with Online World Modeling (OWM) and Adversarial World Modeling (AWM). OWM requires simulator access during 
training and is therefore evaluated only on PushT, PointMaze, and Wall, where fast 
simulators are available. Results from these environments help us establish whether data coverage expansion improves GBP. AWM, which requires no simulator, is evaluated 
on all seven tasks. We report all finetuning, planning, and optimization hyperparameters in \Cref{table:hyperparameters}.

\begin{figure}[!t]
    \centering
    \includegraphics[width=1\linewidth]{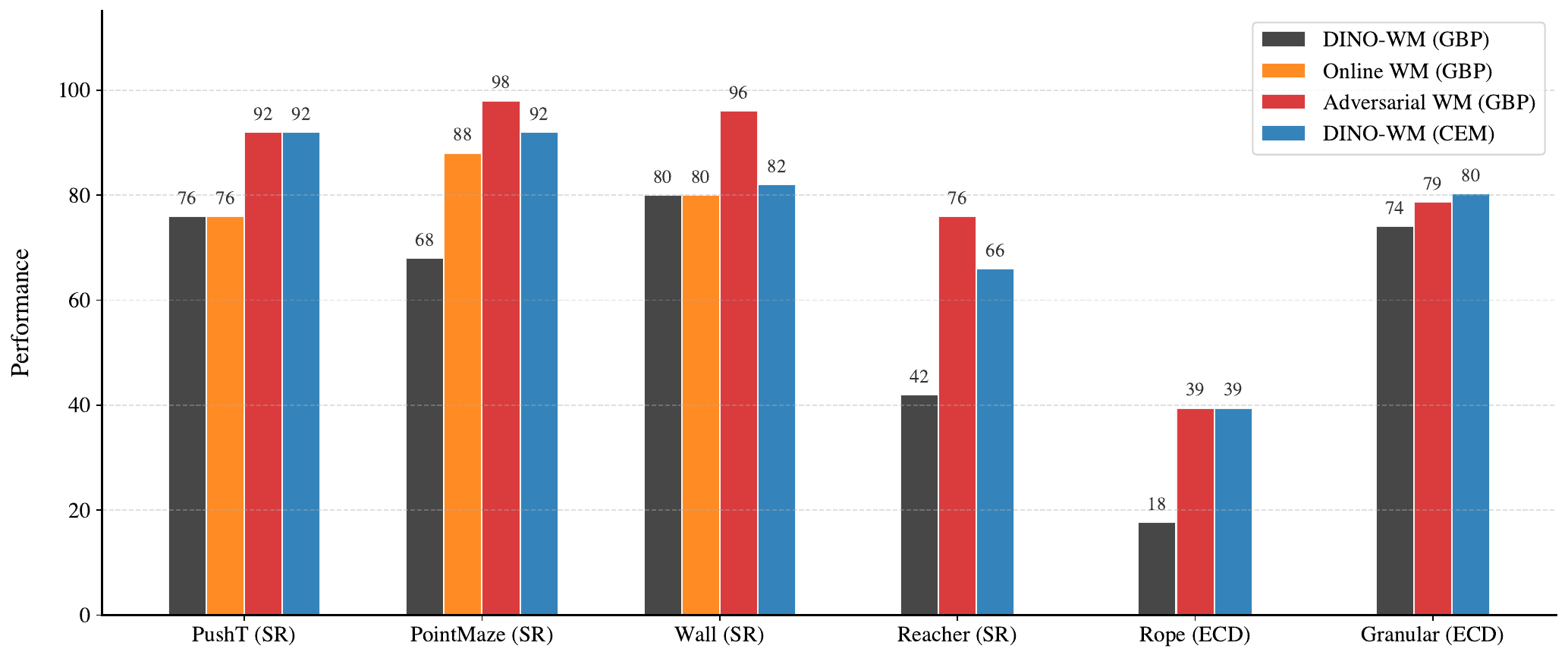}
    \caption{\textbf{Planning Performance on Simulation Tasks.} Comparison of world model planning performance across a variety of domains. For PushT, PointMaze, Wall, and Reacher, we report Success Rate (SR) as a percentage of task completed. For Rope and Granular we report Exponential Chamfer Distance (ECD) \citep{zhang2014robust} which is a measure of point-cloud similarity (larger is better). More details can be found in \Cref{sec:planning_app}.}
    \label{fig:main_results}
\end{figure}

\begin{figure}[!t]
    \centering
    \includegraphics[width=1\linewidth]{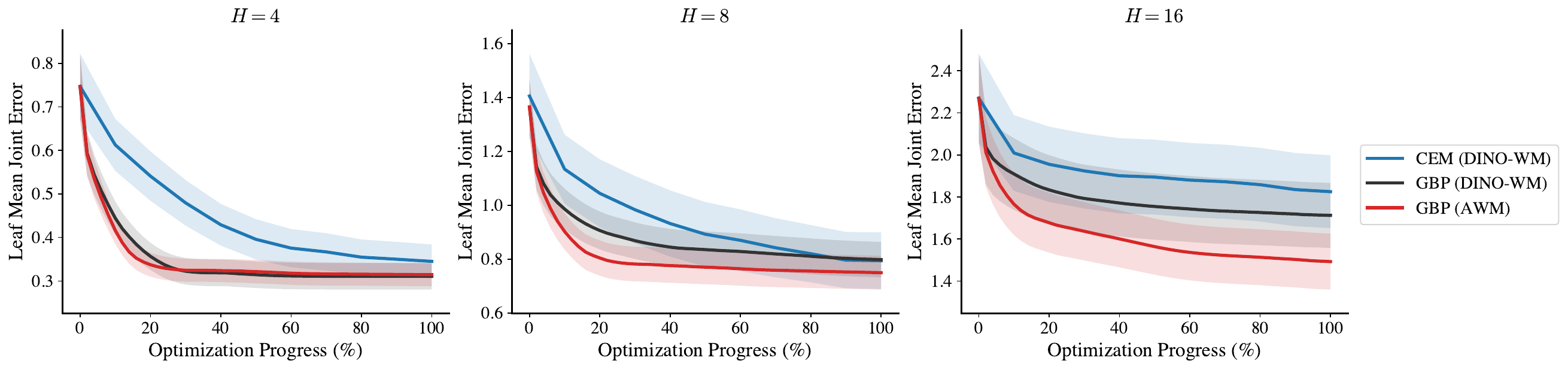}
    \caption{\textbf{Egocentric Body Pose Planning Performance.} We plot the Leaf Mean Joint Error (measure of distance between body poses) over the course of optimization for CEM and GBP with both DINO-WM and AWM. Additional details can be found in \Cref{sec:nymeria}. }
    \label{fig:mje}
\end{figure}

\begin{figure}[!t]
    \centering
    \includegraphics[width=1\linewidth]{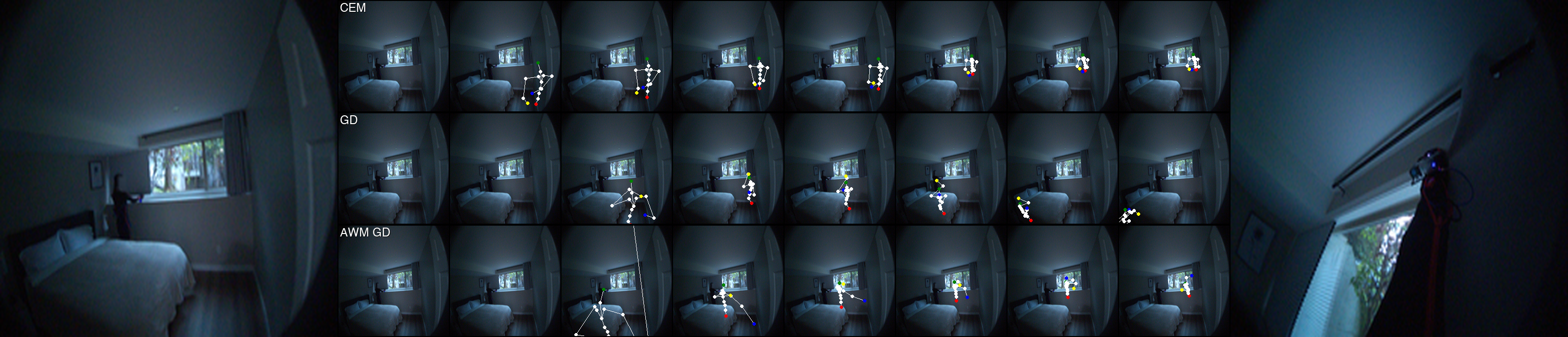}\par\nointerlineskip
    \includegraphics[width=1\linewidth]{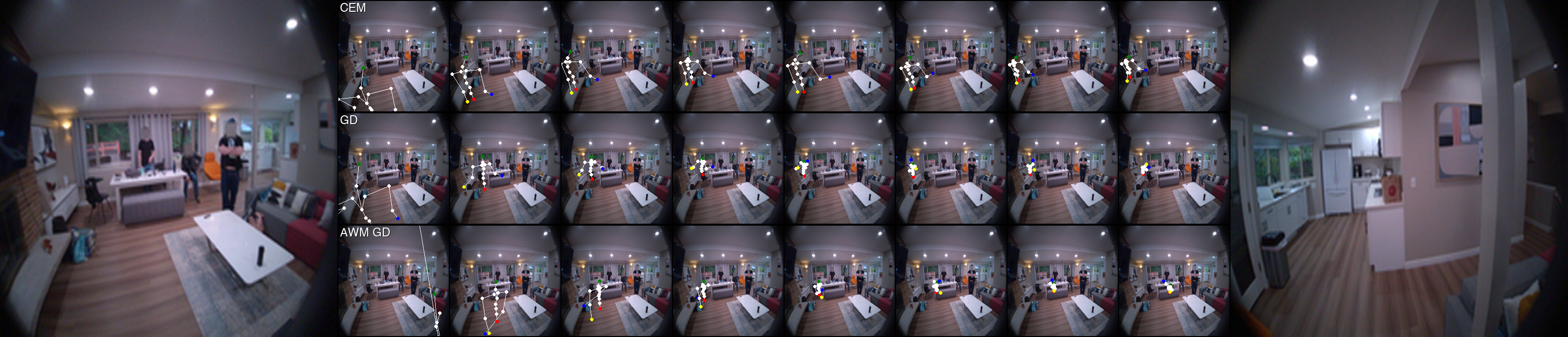}
    
    \caption{\textbf{Visualization of Predicted Body Pose Rollouts.} \textit{Left:} The initial observation. \textit{Middle:} Sequence of 8 future body poses from 16 predicted action with each planning method. \textit{Right:} The goal observation. For visualization, we label the pelvis \textcolor{red}{$\bullet$}, head \textcolor{green}{$\bullet$}, left-hand \textcolor{yellow}{$\bullet$}, and right-hand \textcolor{blue}{$\bullet$}. In the top example, AWM not only reaches the window, but raises its right-hand correctly which CEM fails to do. In the bottom example, we see that AWM is the only method to get close to the gray sofa.}
    \label{fig:planning_visualization}
\end{figure}

\subsection{Planning Results}

As shown in Figure \ref{fig:main_results}, AWM substantially improves over the DINO-WM 
GBP baseline across all six simulated tasks, with gains ranging from $+16\%$ on 
PushT and Wall to $+34\%$ on Reacher. Notably, AWM exceeds CEM on PointMaze 
($98\%$ vs.\ $92\%$), Wall ($96\%$ vs.\ $82\%$), and Reacher ($76\%$ vs.\ $66\%$), 
and matches CEM on PushT ($92\%$) and Rope ($39\%$). On Granular, AWM comes 
within one point of CEM ($79\%$ vs.\ $80\%$).
These results demonstrate that AWM 
enables GBP to consistently match or exceed CEM across a diverse range of 
planning domains, making it a viable method for world model planning.

We surprisingly find that OWM fails to support GBP. On PushT and Wall, OWM produces no improvement over the DINO-WM GBP baseline 
($76\%$ and $80\%$ respectively, identical to the baseline). On PointMaze, OWM 
improves over the baseline ($88\%$ vs.\ $68\%$) but falls short of both AWM 
($98\%$) and CEM ($92\%$). In no environment does OWM match CEM. We analyze the reasons for OWM's poor performance in \Cref{sec:overcoming}.

We also compare AWM to the latent space regularization method LeWM \citep{maes2026leworldmodelstableendtoendjointembedding}. We find that LeWM is unable to perform GBP as effectively as DINO-WM + AWM. Furthermore, finetuning LeWM with AWM improves its GBP performance, demonstrating the versatility of our method. Additional details can be found in \Cref{sec:lewm}.

\subsection{Planning Computational Efficiency}
\begin{figure*}[t]
    \centering
  \includegraphics[width=0.9\linewidth]{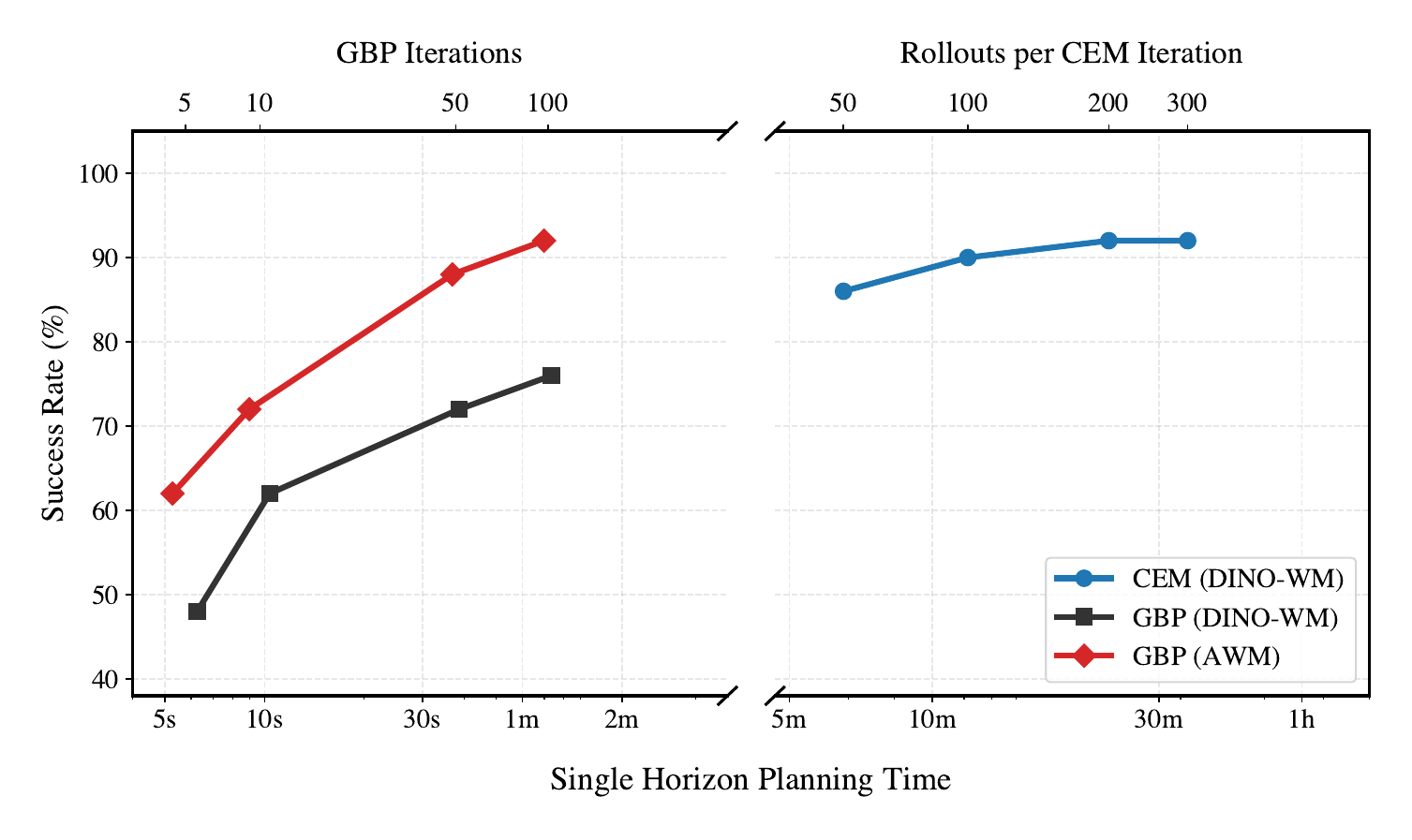}
  \caption{\textbf{Planning efficiency on PushT}. \textit{Left:} GBP success rate 
as a function of wall-clock time controlled by the number of gradient descent iterations taken in a single MPC horizon. \textit{Right:} CEM success rate as a function of wall-clock time 
controlled by the number of rollouts per CEM iteration. Note that the x-axis is logarithmic, highlighting that GBP operates in over an order of magnitude less wall-clock time than CEM.}
  \label{fig:pusht_wall_clock}
\end{figure*}

Deploying world model planning on real robotic systems requires planning that is 
both reliable and fast. While CEM has historically been the dominant planning algorithm because it fulfills the reliability aspect, GBP scales faster by design. Improving performance on GBP requires only more gradient descent iterations over a single 
world model rollout, whereas CEM must sample additional rollouts. This distinction severely compounds. \Cref{fig:pusht_wall_clock} illustrates this directly on PushT. On a single A100 GPU, GBP optimizes a single horizon of actions on the order of seconds to a couple of minutes, while CEM is on the order of a few minutes to over half an hour. AWM with GBP therefore establishes a new Pareto frontier for world model planning: it 
achieves performance that matches or exceeds CEM at over an order of magnitude lower wall-clock cost.



\subsection{Overcoming GBP's failure modes}
\label{sec:overcoming}
We study the effectiveness of OWM to determine if data coverage over latent states that are encountered during planning but might not have been seen during training can lead to gains in closing the model-simulator gap. As we see in \Cref{fig:main_results}, OWM is inconsistent in improving GBP performance. Figure \ref{fig:surrogate_gap} demonstrates that OWM fails to decrease the model-simulator gap, which suggests that data coverage is not necessarily the underlying cause of the gap. The significance of this discovery is that compute and data scaling alone will not make GBP viable, and so we must turn towards the optimization process itself. Figure \ref{fig:landscape} showing OWM does not appear to change the smoothness of the actions optimization landscape only further motivates a need for a method that does. As expected, we see that AWM successfully smooths the loss landscape, getting rid of many local minima. We also see that AWM improves the model-simulator gap, suggesting that a smooth landscape is crucial for GBP to find faithful solutions.

\subsection{Realistic Egocentric Pose Planning}

One of the key disadvantages of CEM is the exponentially large search space that arises from high-dimensional actions. We further highlight the benefits of AWM improving GBP performance by evaluating on the task of future body pose prediction using real motion capture and egocentric video data. Specifically, we use the Nymeria \cite{nymeria24} dataset that features a 48 dimensional action space for control over an entire body pose. The details of the dataset can be found in Appendix \Cref{sec:nymeria}. 

Figure \ref{fig:mje} shows that GBP with AWM significantly improves Leaf Mean Joint Error (MJE) on longer horizons over GBP with DINO-WM. Notably, CEM with DINO-WM underperforms both GBP methods on the longest-horizon, indicating that GBP is more suitable for humanoid-style tasks that require long-horizon planning. Figure \ref{fig:planning_visualization} provides visualizations justifying AWM's superior planning ability that allows it to achieve the most physically plausible goal poses.

\section{Limitations}
We conclude that OWM is an ineffective solution for improving GBP, however, it is important to reiterate that it requires a simulator to improve data coverage, a tool that is not available for realistic tasks. In \Cref{sec:awm-ablations}, we see that setting perturbation radii to be too large can degrade performance, and so AWM introduces new hyper-parameters that might require tuning aside from GBP's hyper-parameters. Although AWM achieves strong performance in long-horizon, realistic environments, it is not yet clear if AWM+GBP could run reliably on real hardware at a reasonable frequency. 

\section{Conclusion}
In this work, we introduce \emph{Online World Modeling} and \emph{Adversarial World Modeling} as techniques for addressing the train-test gap that arises when world models trained on next-state prediction are used for iterative gradient-based planning. Across our experiments, these methods substantially improve the reliability of GBP and, in some settings, allow it to match or outperform sampling-based planners such as CEM. By narrowing this gap, our results suggest that gradient-based planning can be a practical alternative for planning with world models, particularly in settings where computational efficiency is critical. An important direction for future work is to implement these methods on real-world hardware. Adversarial training may additionally improve a world model's robustness to environmental adversaries or stochasticity. More broadly, world models offer a natural advantage over policy-based reinforcement learning in long-horizon decision making. Our work can make significant progress in the development of humanoid robots. Automating physical labor can have massive positive impact in many industries.

\newpage
\bibliographystyle{plainnat}
\bibliography{neurips}


\newpage
\appendix
\section{Related Work}

\textbf{Learning world models from sensory data.}
Learning-based dynamics models have become central to control and decision making, offering a data-driven alternative to classical approaches that rely on first principles modeling~\citep{goldstein1950classical, Schmidt2009DistillingFNA, macchelli2009port}. Early work focused on modeling dynamics in low-dimensional state-space~\citep{deisenroth2011pilco, lenz2015deepmpc, henaff2017model, Sharma2019DynamicsAwareUD}, while more recent methods learn directly from high-dimensional sensory inputs such as images. Pixel-space prediction methods~\citep{Finn2016UnsupervisedLFA, Kaiser2019ModelBasedRL} have shown success in applications such as human motion prediction~\citep{Finn2016UnsupervisedLFA}, robotic manipulation~\citep{Finn2016DeepVF, agrawal2016learning, zhang2019solar}, and solving Atari games~\citep{Kaiser2019ModelBasedRL}, but they remain computationally expensive due to the cost of image reconstructions. To address this, alternative approaches learn a compact latent representation where dynamics are modeled~\citep{Karl2016DeepVB, hafner2019learning, Shi2022RoboCraftLT, karypidis2024dino}. These models are typically supervised either by decoding latent predictions to match ground truth observations~\citep{Edwards2018ImitatingLPA, Zhang2021DeformableLOA, bounou2021online, Hu2022ModelBasedILA, Akan2022StretchBEVSFA, hafner2019learning}, or by using prediction objectives that operate directly in latent space, such as those in joint-embedding prediction architectures (JEPAs)~\citep{lecun2022path, Bardes2024RevisitingFPA, Drozdov2024VideoRLA, Guan2024WorldMFA, zhou2025dinowmworldmodelspretrained}. Our work builds upon this latter category of world models and specifically leverages the DINOv2-based latent world models introduced in~\cite{zhou2025dinowmworldmodelspretrained}. However, unlike prior work that primarily targets improving general representation quality or prediction accuracy, we focus on enhancing the trainability of world models to improve the convergence and reliability of gradient-based planning.

\textbf{Planning with world models.}
Planning with world models is challenging due to the non-linearity and non-convexity of the objective. Search-based methods such as CEM \citep{rubinstein2004cross} and MPPI~\citep{williams2017model} are widely used in this context~\citep{Williams2017InformationTMA, Nagabandi2019DeepDMA, hafner2019learning, Zhan2021ModelBasedOPA, zhou2025dinowmworldmodelspretrained}. These methods explore the action space effectively, helping to escape from local minima, but typically scale poorly in high-dimensional settings due to their sampling-based nature.
In contrast, gradient-based methods offer a more scalable alternative by exploiting the differentiability of the world model to optimize actions directly via backpropagation. Despite their efficiency, these methods suffer from local minima in highly non-smooth loss landscapes~\citep{Bharadhwaj2020ModelPredictiveCVA, Xu2022AcceleratedPLA, Chen2022BenchmarkingDOA, Wang2023SoftZooASA}, and gradient optimization can induce adversarial action sequences that exploit model inaccuracies~\citep{Schiewer2024ExploringTLA, Jackson2024PolicyGuidedDA}. \cite{zhou2025dinowmworldmodelspretrained} have observed that GBP is particularly brittle when used with world models built on pre-trained visual embeddings, such as DINOv2~\citep{oquab2024dinov2learningrobustvisual}, often underperforming compared to CEM. To address these challenges, several stabilizing techniques have been proposed. For instance, random-sampling shooting helps mitigate adversarial trajectories by injecting noise in the action sequence and exploring a broader set of actions during trajectory optimization~\citep{nagabandi2018neural}, and~\cite{Zhang2025StateAwarePOA} introduce adversarial attacks on learned policies to make them robust to environmental perturbations by selectively perturbing state inputs at inference time. In contrast, we apply perturbation directly to latent states and latent actions during world model training.
\cite{florence2022implicit} add gradient penalties when training an implicit policy function to improve its smoothness and stabilize optimization, but their method does not involve training or using a world model.
Other approaches aim to use a hybrid method that combines search and gradient steps to balance global exploration and local refinement~\citep{Bharadhwaj2020ModelPredictiveCVA}. In our work, we modify the world-model training procedure itself to improve GBP stability. In particular, through our Adversarial World Modeling approach, we enhance the robustness of the world model to perturbed states and actions, producing more stable and informative gradients that prevent adversarial action sequences at test time.

\section{AWM Algorithm}
\begin{algorithm}[h]
\caption{Adversarial World Modeling (AWM)}
\label{algo:adv_wm}
\begin{algorithmic}
\STATE {\bfseries Input:} Pretrained world model $f_\theta$, dataset of trajectories $\mathcal{T}$, action perturbation scaling $\lambda_a$, state perturbation scaling $\lambda_z$, horizon $H$, training iterations $N$, minibatch size $B$
\STATE Initialize new trajectory dataset $\mathcal{T}'$
\FOR{$i = 1, \dots, N$}
    \STATE Sample minibatch $\tau \gets \{(z_t^j, a_t^j, z_{t+1}^j)_{t=1}^H\}_{j=1}^{B} \sim \mathcal{T}$
    \STATE $\epsilon_a \gets \lambda_a\, \mathrm{mean}_j\big[\mathrm{std}(\{a_1^j, \ldots, a_H^j\})\big]$
    \STATE $\epsilon_z \gets \lambda_z\, \mathrm{mean}_j\big[\mathrm{std}(\{z_1^j, \ldots, z_{H+1}^j\})\big]$
    \STATE $(\alpha_a, \alpha_z) \gets (1.25\,\epsilon_a, 1.25\,\epsilon_z)$
    \STATE $\delta_a \sim \mathrm{Uniform}(-\epsilon_a, \epsilon_a)$
    \STATE $\delta_z \sim \mathrm{Uniform}(-\epsilon_z, \epsilon_z)$
    \FOR{$t = 1, \dots, H$}
        \STATE $\nabla_{\delta_a}, \nabla_{\delta_z} \gets \nabla_{\delta_a, \delta_z}\big\lVert f_\theta(z_t + \delta_z, a_t + \delta_a) - z_{t+1} \big\rVert_2^2$
        \STATE $\delta_a \gets \mathrm{clip}(\delta_a + \alpha_a \,\mathrm{sign}(\nabla_{\delta_a}), -\epsilon_a, \epsilon_a)$
        \STATE $\delta_z \gets \mathrm{clip}(\delta_z + \alpha_z \,\mathrm{sign}(\nabla_{\delta_z}), -\epsilon_z, \epsilon_z)$
        \STATE $a_t' \gets a_t + \delta_a$
        \STATE $z_t' \gets z_t + \delta_z$
    \ENDFOR
    \STATE $\tau' \gets \{(z_t^{\prime j}, a_t^{\prime j}, z_{t+1}^j)_{t=1}^H\}_{j=1}^{B}$
    \STATE Train $f_\theta$ on next-state prediction using $\tau'$
\ENDFOR
\end{algorithmic}
\end{algorithm}

\section{Experimental Details}
\subsection{Simulation Task Details}
\label{sec:task_details}

\paragraph{\textbf{PushT:}} This task introduced by \cite{pusht} uses an agent interacting with a T-shaped block to guide both the agent and block from a randomly initialized state to a feasible goal state within 25 steps. We use the dataset of 18500 trajectories given in \cite{zhou2025dinowmworldmodelspretrained}, in which the green anchor serves purely as a visual reference.  We draw a goal state from one of the noisy expert trajectories at 25 steps from the starting state.

\paragraph{\textbf{PointMaze:}} In this task introduced by \cite{pointmaze}, a force-actuated ball which can move in the $x,y$ Cartesian directions has to reach a target goal within a maze. We use the dataset of 2000 random trajectories present in \cite{zhou2025dinowmworldmodelspretrained}, with a goal state chosen 25 steps from the starting state.

\paragraph{\textbf{Wall:}} This task introduced by DINO-WM \citep{zhou2025dinowmworldmodelspretrained} features a 2D navigation environment with two rooms separated by a wall with a door. The agent’s task is to navigate from a randomized starting location in one room to a random goal state in the other room, passing through the door. We use the dataset of 1920 trajectories as provided in DINO-WM, with a goal state chosen 25 steps from the starting state.

\paragraph{\textbf{Reacher:}} This is a continuous control task from the DeepMind Control Suite \citep{tassa2018deepmindcontrolsuite}, where a 2-joint robotic arm
aims to reach a target position in a fixed 2D plane. We use a dataset of 10000 trajectories of 200 steps each from \citep{maes2026leworldmodelstableendtoendjointembedding}, that were generated using an expert Soft Actor-Critic policy.

\paragraph{\textbf{Rope:}} In this task introduced by \cite{zhang2024adaptigraph} a simulated Xarm must push a piece of rope into the goal orientation. We use the dataset of 1000 trajectories of 20 steps each provided in DINO-WM.

\paragraph{\textbf{Granular:}} In this task introduced by \cite{zhang2024adaptigraph} a simulated Xarm must push around one hundred small particles into the goal configuration. We use the dataset of 1000 trajectories of 20 steps each provided in DINO-WM.

\begin{figure}[!h]
    \centering
    \begin{subfigure}[t]{0.15\textwidth}
        \centering
        \includegraphics[width=\textwidth]{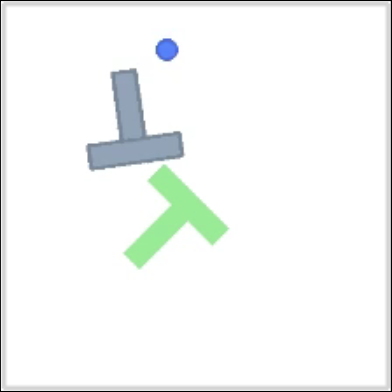}
        \caption*{PushT}
    \end{subfigure}
    \hfill
    \begin{subfigure}[t]{0.15\textwidth}
        \centering
        \includegraphics[width=\textwidth]{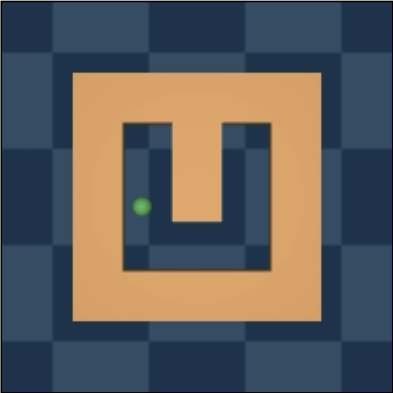}
        \caption*{PointMaze}
    \end{subfigure}
    \hfill
    \begin{subfigure}[t]{0.15\textwidth}
        \centering
        \includegraphics[width=\textwidth]{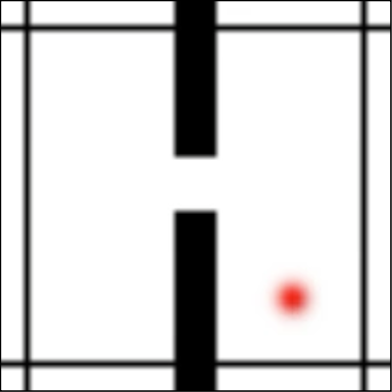}
        \caption*{Wall}
    \end{subfigure}
    \hfill
    \begin{subfigure}[t]{0.15\textwidth}
        \centering
        \includegraphics[width=\textwidth]{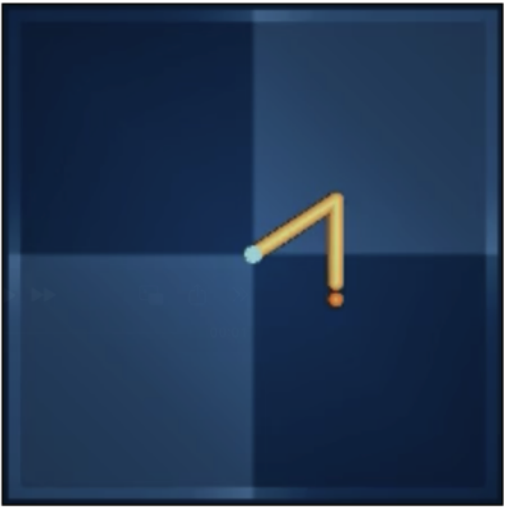}
        \caption*{Reacher}
    \end{subfigure}
    \hfill
    \begin{subfigure}[t]{0.15\textwidth}
        \centering
        \includegraphics[width=\textwidth]{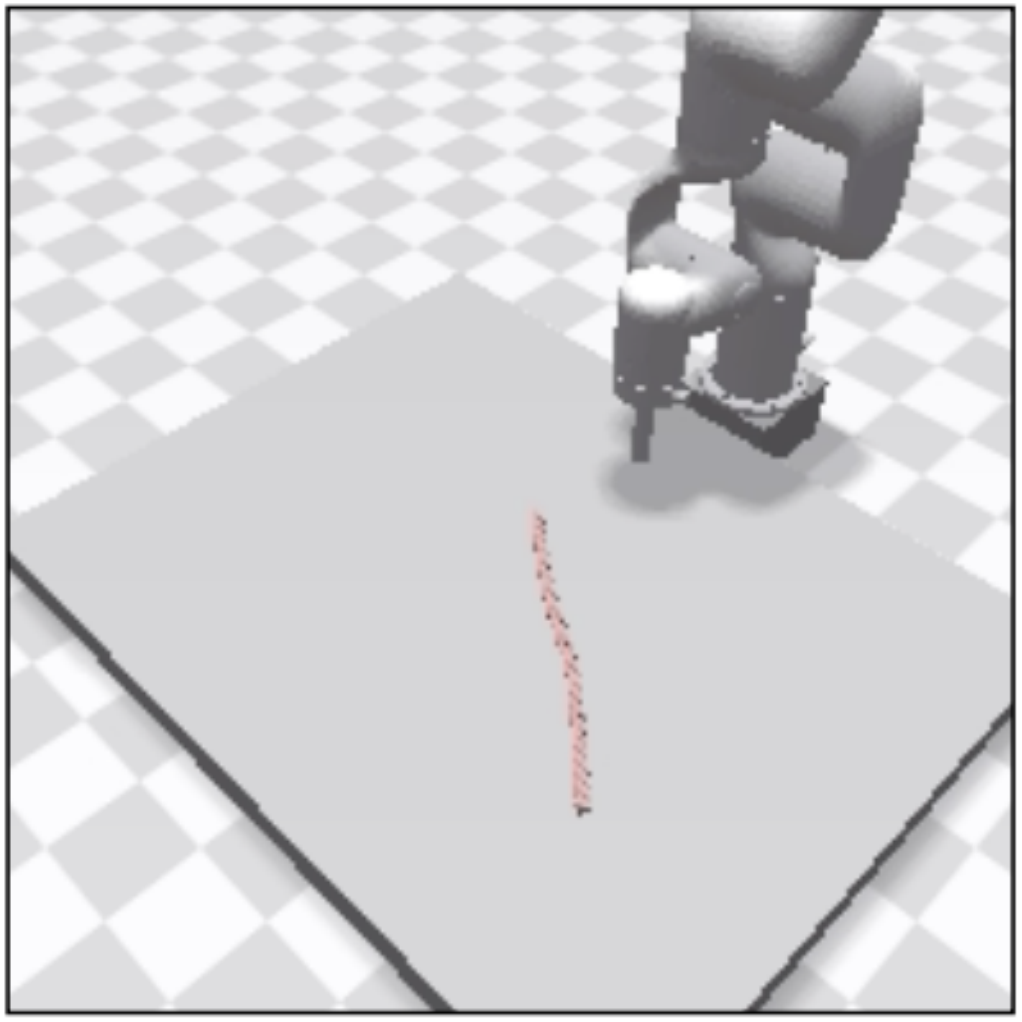}
        \caption*{Rope}
    \end{subfigure}
    \hfill
    \begin{subfigure}[t]{0.15\textwidth}
        \centering
        \includegraphics[width=\textwidth]{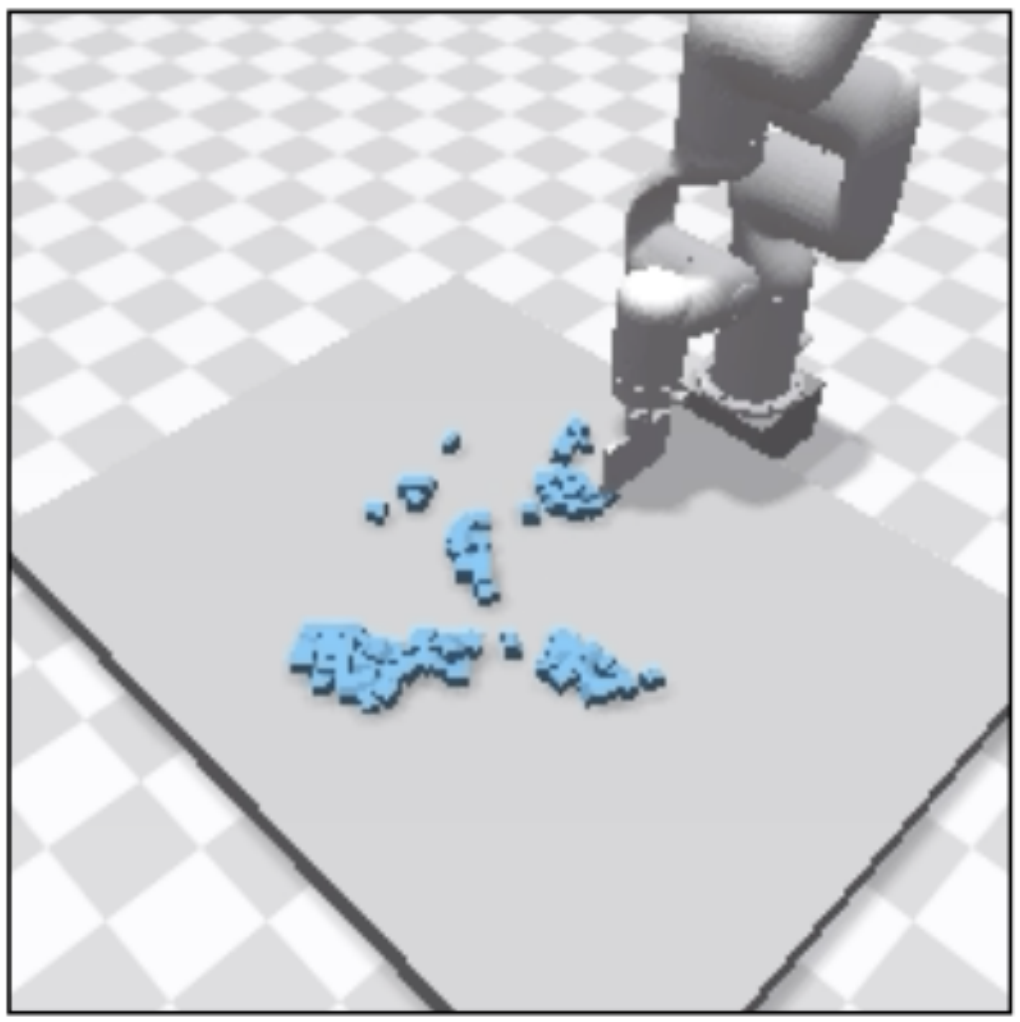}
        \caption*{Granular}
    \end{subfigure}

    \caption{Illustrations of the six simulation tasks used in our experiments. Images from \cite{zhou2025dinowmworldmodelspretrained}.}
    \label{fig:tasks}
\end{figure}

Each of the tasks' environment is illustrated in \Cref{fig:tasks}. We reproduce the dataset statistics used to train the base world model for each environment from \cite{zhou2025dinowmworldmodelspretrained} as detailed in \Cref{tab:dataset_info}.
\begin{table}[!h]
    \caption{Trajectory datasets used to pretrain the base DINO-WM.}
    \centering
    \begin{tabular}{c|cccc}
        \toprule
       Environment & History & Frameskip & Dataset Size & Trajectory Length \\
       \midrule
       Push-T  & 3 & 5 & 18500 & 100-300 \\
       PointMaze & 3 & 5 & 2000 & 100  \\
       Wall & 1 & 5 & 1920 & 50 \\
       Reacher & 3 & 5 & 10000 & 200 \\
       Rope & 1 & 1 & 1000 & 5 \\
       Granular & 1 & 1 & 1000 & 5 \\
       \bottomrule
    \end{tabular}
    
    \label{tab:dataset_info}
\end{table}

\subsection{CEM Algorithm}
\label{sec:cem_description}
We detail the cross-entropy method used in our planning experiments in \Cref{algo:cem}.

\begin{algorithm}[tb]
\caption{Cross-Entropy Method (CEM) Planning}
\label{algo:cem}
\begin{algorithmic}
\STATE {\bfseries Input:} Current observation $o_0$, goal observation $o_g$, encoder $\Phi_\mu$, world model $f_\theta$, horizon $H$, population size $N$, top-K selection $K$, iterations $I$
\STATE $\hat{z}_1 \leftarrow \Phi_\mu(o_1)$
\STATE $z_g \leftarrow \Phi_\mu(o_g)$
\STATE Initialize Gaussian distribution parameters: mean $\mu_0$, covariance $\Sigma_0$
\FOR{$i = 1, \dots, I$}
  \STATE Sample $N$ action sequences $\{a^{(j)}_{1:H}\}_{j=1}^N \sim \mathcal{N}(\mu_{i-1}, \Sigma_{i-1})$
  \FOR{$j = 1, \dots, N$}
    \STATE $\hat{z}^{(j)}_1 \leftarrow \hat{z}_1$
    \FOR{$t = 2, \dots, H+1$}
      \STATE $\hat{z}^{(j)}_t \leftarrow f_\theta(\hat{z}^{(j)}_{t-1}, a^{(j)}_{t-1})$
    \ENDFOR
    \STATE $C^{(j)} \leftarrow \lVert \hat{z}^{(j)}_{H+1} - z_g \rVert^2$
  \ENDFOR
  \STATE Select $K$ sequences with lowest cost: $\mathcal{E} = \{a^{(j)}\}_{\text{top-}K}$
  \STATE $\mu_i \leftarrow \frac{1}{K} \sum_{a \in \mathcal{E}} a$
  \STATE $\Sigma_i \leftarrow \frac{1}{K} \sum_{a \in \mathcal{E}} (a - \mu_i)(a - \mu_i)^\top$
\ENDFOR
\STATE $\textbf{return } \mu_I \text{ as the final action sequence estimate} \{\hat{a}_t\}_{t=1}^{H}$ \;

\end{algorithmic}
\end{algorithm}

\subsection{Finetuning and Planning Hyperparameters}
\label{sec:hyperparameters}

In \Cref{table:hyperparameters}, we list all shared hyperparameters used in training and planning. For all planning experiments we use 10 MPC iterations, taking the first $K = 1$ actions at each iteration before replanning. For all simulator tasks, we use a horizon of $H = 5$ actions. We provide data quantity and synthetic data parameters for our Online and Adversarial World Modeling training setups in \Cref{table:awm_train_params,table:owm_train_params} respectively. In addition to maintaining perturbation radii for the visual latent and action embeddings, we use a distinct radius for the proprioceptive embeddings. We empirically find that the scales of the visual and proprioceptive embeddings are incompatible and semantically distinct, thereby necessitating independent perturbation. Throughout all of our experiments, we set the perturbation radii of the action embedding and proprioceptive embedding identically for simplicity.

\begin{table}[!h]
\caption{We report \textbf{(a)} shared hyperparameters for OWM/AWM finetuning across all environments, \textbf{(b)} GBP parameters, and  \textbf{(c)} CEM parameters.}
\centering
\captionsetup[subtable]{justification=centering}

\begin{subtable}{0.30\linewidth}
\caption{Finetuning Parameters}
\centering
\begin{tabular}{lc}
    \toprule
    Name & Value  \\
    \midrule
    Image size & 224 \\
    Optimizer & AdamW \\
    Predictor LR & 1e-5 \\
    \bottomrule
\end{tabular}

\end{subtable}
\hfill
\begin{subtable}{0.30\linewidth}
\caption{GBP}
\centering
\begin{tabular}{lcc}
    \toprule
    Name & Value  \\
    \midrule
    Optimizer steps & 100 \\
    LR & 0.2 \\
    Optimizer & Adam \\
    \bottomrule
\end{tabular}

\end{subtable}
\hfill
\begin{subtable}{0.30\linewidth}
\caption{CEM}
\centering
\begin{tabular}{lcc}
    \toprule
    Name & Value \\
    \midrule
    Iterations        & 30 \\
    Number of Elites & 30 \\
    Rollouts per Iteration   & 300 \\
    \bottomrule
\end{tabular}

\end{subtable}

\label{table:hyperparameters}
\end{table}

\begin{table}[!h]
\caption{Finetuning parameters for Online World Modeling.}
\label{table:owm_train_params}
\centering
\begin{tabular}{r|c|c|c|c}
\toprule
\textbf{Environment} & \textbf{\# Rollouts} & \textbf{Batch Size} & \textbf{GPU} &  \textbf{Epochs} \\
\midrule
PushT       & 6000 & 32 & 4x B200 & 1 \\
PointMaze   & 500 & 32 & 4x B200 & 1 \\
Wall        & 1920 (all) & 80 & 4x B200 & 1 \\
\bottomrule
\end{tabular}

\end{table}

\begin{table}[!h]
\caption{Training parameters for Adversarial World Modeling.}
\label{table:awm_train_params}
\centering
\begin{tabular}{r|c|c|c|c|c|c|c}
\toprule
\textbf{Environment} & \textbf{\# Rollouts} & \textbf{Batch Size} & \textbf{GPU} & \textbf{Epochs} & \textbf{$\eps_\text{visual}$} & \textbf{$\eps_\text{proprio}$} & \textbf{$\eps_\text{action}$} \\
\midrule
PushT       & 20000 (all) & 16 & 8x B200 & 1   & 0.05            & 0.02                    & 0.02            \\
PointMaze   & 2000 (all) & 16 & 1x B200  & 1  & 0.20            & 0.08                    & 0.08            \\
Wall        & 1920 (all) & 48 & 1x B200 & 2    & 0.20            & 0.08                    & 0.08           \\
Reacher       & 10000 (all) & 32 & 1xA100 & 2   & 0.05            & 0.02                    & 0.02            \\
Rope       & 1000 (all) & 16 & 1xRTX3080 & 1   & 0.05            & 0.02                    & 0.02            \\
Granular       & 1000 (all) & 16 & 1xRTX3080 & 1   & 0.05            & 0.02                    & 0.02            \\
\bottomrule
\end{tabular}

\end{table}

\subsection{Weighted Goal Loss}
\label{sec:wgl}
We use an exponentially weighted goal loss to improve progress in navigation-based tasks. Weighted Goal Loss (WGL) minimizes the distance between all latent states in the predicted rollout and the goal latent state, but weighted by an exponential weighting. Formally,
\begin{equation}
    \mathcal{L}_\text{WGL} = \frac{1}{H}\sum_{i = 2}^{H + 1} w_{i}\lVert\hat{z}_i - z_{\text{goal}} \rVert^2_2.
\end{equation}
where the sequence of normalized weights $\{w_i\}_2^{H+1}$ is a hyperparameter choice. Empirically, we find that using this objective for Gradient-Based Planning either maintains or improves planning performance. For PointMaze and Wall, we found that exponentially upweighting later states in the planning horizon improved planning performance, so we set $w_i = 2^i$. For PushT, we found that exponentially upweighting earlier states improved planning performance, so we set $w_i = \left(1/2\right)^i$. We leave the optimal selection of this sequence of weights as future work.

\subsection{Simulation Tasks Planning Results}
\label{sec:planning_app}
In Table \ref{table:planning_table}, we report mean planning performance across a total of 50 trajectories for each method and task. We calculate standard deviations using the averages of 3 batches of size 16. Exponential Chamfer Distance \cite{zhang2014robust} is a bounded version (from 0 - 100) of the traditional chamfer distance, allowing us to measure the performance of Rope and Granular on the same scale as the other simulation tasks.

\begin{table}[!h]
\caption{Planning performance across simulation tasks and methods.}
\label{table:planning_table}
\centering
\begin{tabular}{r|c|c|c|c}
\toprule
\textbf{Environment} & \textbf{DINO-WM (GBP)} & \textbf{AWM (GBP)} & \textbf{OWM (GBP)} & \textbf{DINO-WM (CEM)} \\
\midrule
PushT (SR)     & $76 \pm 9.31$   & $92 \pm 5.55$   & $76 \pm 8$      & $92 \pm 2.69$   \\
PointMaze (SR) & $68 \pm 10.52$  & $98 \pm 2.77$   & $88 \pm 4.65$   & $92 \pm 5.73$   \\
Wall (SR)      & $80 \pm 12$     & $96 \pm 2.86$   & $80 \pm 11.79$  & $82 \pm 9.62$   \\
Reacher (SR)   & $42 \pm 8.87$   & $76 \pm 4$      & -               & $66 \pm 7.47$   \\
Granular (ECD)  & $74.08 \pm 2.96$ & $78.66 \pm 2.36$ & -               & $80.25 \pm 2.41$ \\
Rope (ECD)     & $17.73 \pm 3.19$ & $39.46 \pm 4.34$ & -               & $39.46 \pm 6.71$ \\
\bottomrule
\end{tabular}
\end{table}

\section{Visualizing the Optimization Landscape}
\label{sec:visualization}
We visualize the loss landscape of both the DINO World Model before and after applying the AWM and OWM objective. We perform a grid search over the subspace spanned by
\begin{enumerate}
    \item $\hat{a}_\text{GBP-Pretrained}$: Gradient-Based Planning on original Dino World Model with 300 optimization steps of Adam with LR = 1e-3. We set a fixed initialization $a_\text{init}$.
    \item $\hat{a}_\text{GBP-Adversarial}$: Gradient-Based Planning on our Adversarial World Model (or Online World Model) with 300 optimization steps of Adam with LR = 1e-3. We use the same $a_\text{init}$ as our initialization.
    \item $a_\text{GT}$: the ground-truth actions from the expert demonstrator.
\end{enumerate}

We define the axes as $\alpha = \hat{a}_{\text{GBP-Pretrained}} - a_{\text{GT}}$ and $\beta = \hat{a}_{\text{GBP-Adversarial}} - a_{\text{GT}}$, and compute the loss surface over a $50 \times 50$ grid spanning $\alpha, \beta \in [-1.25, 1.25]$.

The ``Simulator'' landscape is obtained by executing all interpolated action sequences in the ground-truth environment and computing the embedding-space loss; it validates that the basin near the origin is indeed low-loss and that the overall landscape shape is consistent across models. Since the interpolated actions become physically implausible far from the origin, the simulator landscape exhibits edge artifacts at extreme $(\alpha, \beta)$ values. We reduce opacity beyond $|\alpha|, |\beta| > 0.75$ in the 3D surface plot for clarity.

\section{Adversarial World Modeling: Design Decisions}
\label{sec:awm-design}

\subsection{Fast Gradient Sign Method (FGSM) vs. Projected Gradient Descent (PGD)}
\label{sec:fgsm-vs-pgd}
Projected Gradient Descent (PGD) has been used as an iterative method for generating adversarial perturbations \citep{madry2019deeplearningmodelsresistant}. At each step, PGD takes a gradient ascent step and projects the result onto the space of allowed perturbations (some ball with radius $\eps$ around the input). Projection ($\Pi$) is typically via clipping or scaling. Formally,
\begin{equation}
    \delta^{(k+1)} = \Pi_{\lVert \delta \rVert_\infty \le \epsilon} \left( \delta^{(k)} + \alpha \cdot \nabla_x \mathcal{L}(f_\theta(x + \delta^{(k)}), y) \right)
\end{equation}

However, this is computationally expensive to use for adversarial training as it requires an additional backward pass for each iteration. If one uses a single-step, replaces the gradient by its sign, and uses step size $\alpha = \eps$, this recovers the well-known Fast Gradient Sign Method (FGSM) update \citep{goodfellow2014explaining}.
\begin{equation}
    \delta = \eps \text{sign}\left(\nabla_x \mathcal{L}(f_\theta(x), y)\right).
\end{equation}

In \citealt{fastbetterthanfree}, the authors demonstrate that initializing $\delta$ in the $\ell_\infty$-ball with radius $\eps$ and performing FGSM adversarial training on these perturbations substantially improves robustness to PGD attacks and matches performance of PGD-based training. We leverage this observation to perform cheap adversarial training that only requires $2\times$ the backward passes of traditional supervised learning. In comparison, $K$-step PGD requires $K$ more backward passes ($3\times$ more for $K=2$ and $4\times$ for $K=3$). In \Cref{table:pgd}, we show that 2/3-Step PGD does not consistently outperform FGSM, despite requiring a much larger training budget.

\begin{table}[!b]
\caption{Comparison between 2/3-Step PGD and FGSM demonstrating that FGSM has improved or only slightly subpar performance for significantly less training budget.}
\centering
\setlength{\tabcolsep}{6pt} 
\renewcommand{\arraystretch}{1.1} 

\begin{tabular}{l c cc cc} 
\toprule
& & \multicolumn{2}{c}{\textbf{PointMaze}} & \multicolumn{2}{c}{\textbf{Wall}} \\
\cmidrule(lr){3-4} \cmidrule(lr){5-6}
& \textbf{Backward Passes} & \textbf{Min/Epoch} & \textbf{GBP SR(\%)} & \textbf{{Min/Epoch}} & \textbf{GBP SR(\%)} \\
\midrule
FGSM        & \textbf{2} & \textbf{120} & 94          & \textbf{14} & \textbf{94} \\
2-Step PGD  & 3          & 165          & \textbf{96} & 20          & 90          \\
3-Step PGD  & 4          & 201          & 94          & 24          & 94          \\
\bottomrule
\end{tabular}
\label{table:pgd}
\end{table}

\subsection{Scaling Factor ($\lambda$) \& Perturbation Radii ($\epsilon$) Ablations}
\label{sec:awm-ablations}
\begin{figure}[!h]
    \centering
    \includegraphics[width=\textwidth]{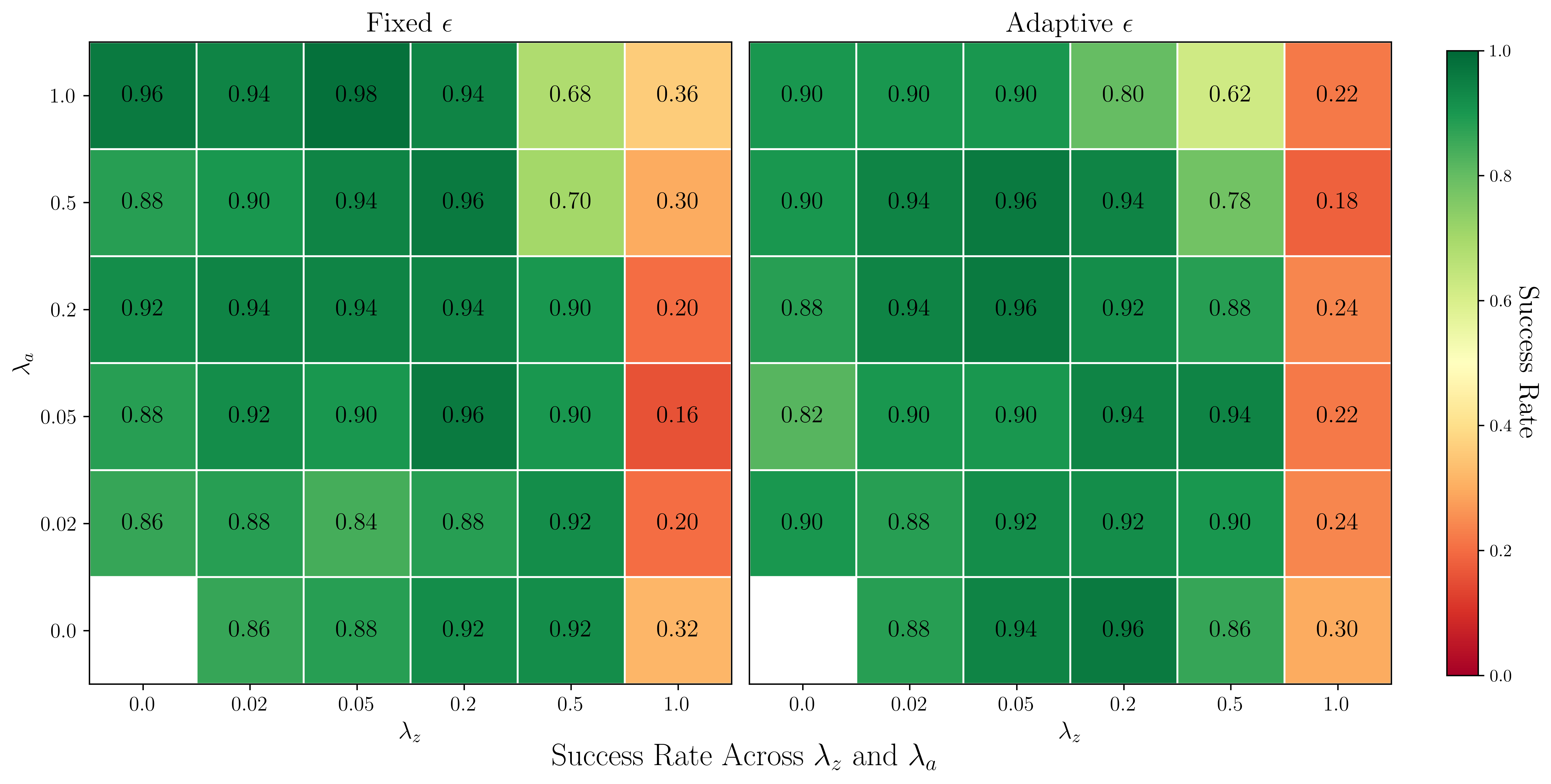}
    \caption{Success rate of GBP planning on an Adversarial World Model trained with scaling factors $\lambda_a, \lambda_z$ and perturbation radii $\eps_a, \eps_z$ on the Wall environment. We find that $0 \le \lambda_z, \lambda_a \le 0.2$ are stable for either ``Fixed'' or ``Adaptive'' perturbation radii.}
    \label{fig:awm_eps_ablation}
\end{figure}
To assess the robustness of Adversarial World Modeling to the scaling factor and perturbation radius hyperparameters, we conduct an ablation study varying these two factors, shown in \Cref{fig:awm_eps_ablation}. We evaluate $\lambda_a, \lambda_z \in [0.0, 0.02, 0.05, 0.20, 0.50, 1.0]^2$ and either fix $\eps_a, \eps_p, \eps_z$ to the standard deviation of the first minibatch (``Fixed") or recompute it for every minibatch (``Adaptive"). We observe no consistent improvement or degradation across any value of $\lambda_a$, for $0 \le \lambda_z \le 0.5$, or between the ``Fixed" or ``Adaptive" perturbation radii. We note that setting the visual scaling factor $\lambda_z$ too high (e.g., $0.5, 1.0$) can significantly degrade performance. We hypothesize that excessively large perturbations distort the semantic content of the visual latent state, pushing it outside the range of semantically equivalent representations.

\section{Egocentric Body Pose Planning and Nymeria Dataset Details}
\label{sec:nymeria}

Nymeria \cite{nymeria24} is a large scale dataset of first-person videos and associated motion capture. The motion capture at a particular timestep is represented by a pose $p_t$, which consists of the 3D position of the pelvis and Euler angles for each joint. An action is defined as the change between two subsequent poses, namely the change in pelvis position and the Euler angles. With 15 joints, the action becomes 48 dimensional. To measure the quality of predicted actions we use Leaf Mean Joint Error (MJE). MJE is defined as
$$
\text{MJE} = \frac{1}{|\text{joints}|}\sum_{j \in \text{joints}} |\hat{x}_j - x_j|.
$$
``Leaf'' indicates that $\hat{x}$ only comes from the final predicted pose. 

We train DINO-WM on 2 epochs of a split containing half the Nymeria dataset, and perform AWM finetuning for 1 epoch on the same split. For CEM planning, we use 128 rollouts per iteration, 16 elites, and 10 iterations. For GBP planning, we use Adam, LR of 0.2, and 50 optimizer steps. 

\section{Comparison to LeWorldModel}
\label{sec:lewm}
LeWorldModel (LeWM) \cite{maes2026leworldmodelstableendtoendjointembedding} is a latent world model that uses a SIGReg \cite{balestriero2025lejepaprovablescalableselfsupervised} regularization term to prevent trivial collapse by enforcing Gaussian-distributed latent states. In \Cref{table:lewm} we show how LeWM alone underperforms DINO-WM with AWM, and that LeWM benefits from AWM as well. For each environment, we finetune LeWM with a single epoch of AWM.

\begin{table}[H]
\caption{GBP performance on PushT and Wall of LeWM and DINO-WM with and without AWM. The same planning setup is used for each architecture.}
\centering
\centering
\begin{tabular}{r|c|c|c|c}
\toprule
\textbf{Environment} & \textbf{DINO-WM} & \textbf{LeWM} & \textbf{LeWM + AWM} & \textbf{DINO-WM + AWM} \\
\midrule
PushT (SR)     & $76 \pm 9.31$   &  $72 \pm 6.33$  & $82 \pm 4.68$      & $92 \pm 5.55$   \\
Wall (SR)      & $80 \pm 12.02$     & $74 \pm 4.53$   & $80 \pm 2.37$  & $96 \pm 2.86$   \\
\bottomrule
\end{tabular}
\label{table:lewm}
\end{table}


\end{document}